\documentclass{article}

\usepackage{arxiv}

\usepackage[utf8]{inputenc} 
\usepackage[T1]{fontenc}    
\usepackage{hyperref}       
\usepackage{url}            
\usepackage{booktabs}       
\usepackage{amsfonts}       
\usepackage{nicefrac}       
\usepackage{microtype}      
\usepackage{lipsum}
\usepackage{graphicx}
\graphicspath{ {./images/} }

\usepackage{multirow}
\usepackage{cite}
\usepackage{hyperref}
\usepackage{booktabs}
\usepackage[normalem]{ulem}
\useunder{\uline}{\ul}{}
\usepackage{rotfloat}
\floatstyle{boxed} 
\restylefloat{figure}

\usepackage{amsmath}

\title{Chatbot Interaction with Artificial Intelligence: Human Data Augmentation with T5 and Language Transformer Ensemble for Text Classification}

\author{
    Jordan J. Bird\\
    ARVIS Lab \\ 
    Aston University, United Kingdom \\
    \texttt{birdj1@aston.ac.uk}
  \And
    Anik\'o Ek\'art \\
    School of Engineering and Applied Science \\ 
    Aston University, United Kingdom \\
    \texttt{a.ekart@aston.ac.uk}
\And
    Diego R. Faria\\
    ARVIS Lab \\ 
    Aston University, United Kingdom \\
    \texttt{d.faria@aston.ac.uk}
}

\begin{document}
\maketitle
\begin{abstract}
In this work, we present the Chatbot Interaction with Artificial Intelligence (CI-AI) framework as an approach to the training of deep learning chatbots for task classification. The intelligent system augments human-sourced data via artificial paraphrasing in order to generate a large set of training data for further classical, attention, and language transformation-based learning approaches for Natural Language Processing. Human beings are asked to paraphrase commands and questions for task identification for further execution of a machine. The commands and questions are split into training and validation sets. A total of 483 responses were recorded. Secondly, the training set is paraphrased by the T5 model in order to augment it with further data. Seven state-of-the-art transformer-based text classification algorithms (BERT, DistilBERT, RoBERTa, DistilRoBERTa, XLM, XLM-RoBERTa, and XLNet) are benchmarked for both sets after fine-tuning on the training data for two epochs. We find that all models are improved when training data is augmented by the T5 model, with an average increase of classification accuracy by 4.01\%. The best result was the RoBERTa model trained on T5 augmented data which achieved 98.96\% classification accuracy. Finally, we found that an ensemble of the five best-performing transformer models via Logistic Regression of output label predictions led to an accuracy of 99.59\% on the dataset of human responses. A highly-performing model allows the intelligent system to interpret human commands at the social-interaction level through a chatbot-like interface (e.g. "Robot, can we have a conversation?") and allows for better accessibility to AI by non-technical users.
\end{abstract}


\section{Introduction}
Attention-based and transformer language models are a rapidly growing field of study within machine learning and artificial intelligence and for applications beyond. The field of Natural Language Processing has especially been advanced through transformers due to their approach to reading being more akin to human behaviour than classical sequential techniques. With many industries turning to Artificially Intelligent solutions by the day, models have a growing requirement for robustness, explainability, and accessibility since AI solutions are becoming more and more popular for those without specific technical backgrounds in the field. Another interesting field that is similarly being seen more often is that of Data Augmentation; that is, creating data from a set that in itself increases the quality of that set of data. The alternative to data augmentation, which is unfortunately the case with many modern NLP systems, is to gather more data. As an alternative to unwanted privacy concerns, data scientists may instead find ways to augment the data as a friendlier alternative. 

\begin{figure}
    \centering
    \includegraphics[scale=1.1]{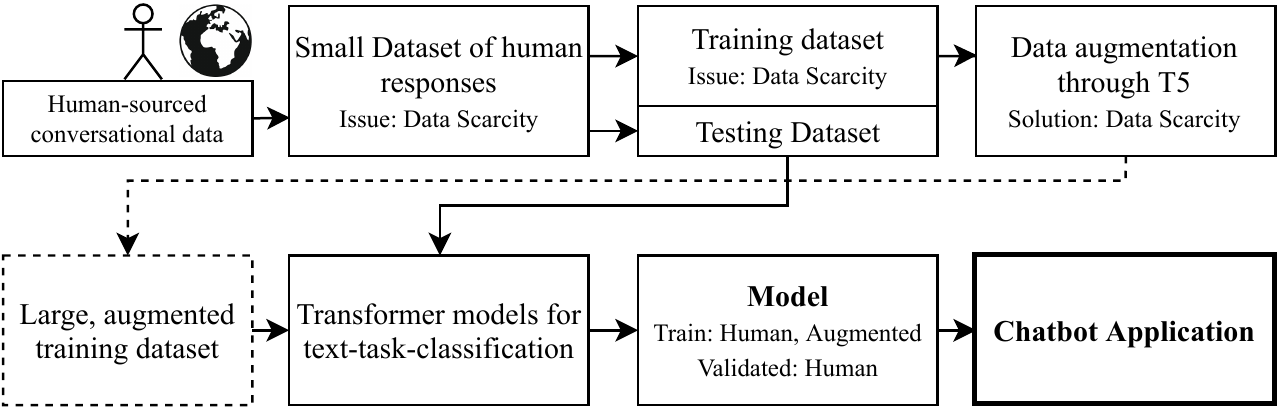}
    \caption{A general overview of the proposed approach.}
    \label{fig:mini-diagram}
\end{figure}

In this study, we bring together all of these aforementioned concepts and fields of study to form a system that we call Chatbot Interaction with Artificial Intelligence (CI-AI). A general overview of the approach can be observed in Figure \ref{fig:mini-diagram}. As an alternative to writing code and managing data, complex machine learning tasks such as conversational AI, sentiment analysis, scene recognition, brainwave classification and sign language recognition among others are given accessibility through an interface of natural, social interaction via both verbal and non-verbal communication. That is, for example, a spoken command of "can we have a conversation?" or a sign language command of "can-we-talk" would command the system to launch a conversational AI program. For such a system to be possible, it needs to be robust, since an interactive system that makes one mistake for many successes would be considered a broken system. The system needs to be accessible to a great number of people with differing backgrounds, and thus must have the ability to generalise by being exposed to a large amount of training data. Last, but by no means least, the system needs to be explainable; as given in a later example, if a human were to utter the phrase, "Feeling sad today. Can you cheer me up with a joke?", which features within that phrase lead to a correct classification and command to the chatbot to tell a joke? Where does the model focus within the given text in order to correctly predict and fulfil the human's request? Thus, to achieve these goals, the scientific contributions of this work are as follows:
\begin{enumerate}
    \item The collection of a 7-class command-to-task dataset from multiple human beings from around the world, giving a total of 483 data objects.
    \item Augmentation of the human data with a transformer-based paraphrasing model which results in a final training dataset of 13,090 labelled data objects.
    \item Benchmarking of 7 State-of-the-Art transformer-based classification approaches for text-to-task commands. Each model is trained on the real training data and validation data, and is then trained on the real training data plus the paraphrased augmented data and validation data. We find that all 7 models are improved significantly when exposed to augmented data.
    \item A deep exploration of the best model. Firstly in order to discern the small amount of errors (1.04\% errors) and how they were caused by seeing the largest errors in terms of loss and the class probability distributions. Secondly, the chatbot is given commands that were not present during training or validation, and top features (words) are observed - interestingly, given their technical nature, the models focus keenly on varying parts of the sentence similar to a human reading. 
\end{enumerate}

The rest of this article is structured as follows. Initially, the background and related studies are explored in Section \ref{sec:background}. The method of the experiments are described in Section \ref{sec:method}, and the results from the experiments are then presented in Section \ref{sec:results}. With the best-performing model in mind, Section \ref{sec:model-exploration} then explores the model in terms of the small number of errors made, and how the model interprets new and unseen data (ie. should the model be in deployment). Finally, conclusions are drawn and future work is suggested in Section \ref{sec:conclusionfuturework}.

\section{Background and Related Works}
\label{sec:background}
The Transformer is a new concept in the field of deep learning~\cite{vaswani2017attention}. Transformers currently have a primary focus on NLP, but state-of-the-art image processing using similar networks have recently been explored~\cite{qi2020imagebert}. With the idea of \textit{paying attention} in mind, the theory behind the exploration of Transformers in NLP is their more natural approach to sentences; rather than focusing on one token at a time in the order that they appear and suffering from the vanishing gradient problem~\cite{schmidhuber1992learning}, Transformer-based models instead pay attention to tokens in a learned order and as such enable more parallelisation while improving upon many NLP problems through which many benchmarks have been broken~\cite{vaswani2017attention,wang2018glue}. For these reasons, such approaches are rapidly forming State-of-the Art scores for many NLP problems~\cite{tenney2019bert}. For text data in particular these include generation~\cite{devlin2018open,radford2019language}, question answering~\cite{shao2019transformer,lukovnikov2019pretrained}, sentiment analysis~\cite{naseem2020transformer,javdan2020applying}, translation~\cite{zhang2018improving,wang2019learning,di2019adapting}, paraphrasing~\cite{chada2020simultaneous,lewis2020pre}, and classification~\cite{sun2019fine,chang2019x}.

According to ~\cite{vaswani2017attention}, Transformers are based on calculation of scaled dot-product attention units. These weights are calculated for each word within the input vector of words (document or sentence). The output of the attention unit are embeddings for a combination of relevant tokens within the input sequence. This is shown later on in Section \ref{sec:model-exploration} where both correctly and incorrectly classified input sequences are highlighted with top features that lead to such a prediction. Weights for the query $W_{q}$, key $W_{k}$, and value $W_{v}$ are calculated as follows:
\begin{equation}
    Attention(Q,K,V) = softmax \left(  \frac{QK^T}{  \sqrt{d_{k}}  }  \right) V.
\end{equation}
The query is an object within the sequence, the keys are vector representations of said input sequence, and the values are produced given the query against keys. Unsupervised models receive $Q$, $K$ and $V$ from the same source and thus pay \textit{self-attention}. For tasks such as classification and translation, $K$ and $V$ are derived from the source and $Q$ is derived from the target. For example, $Q$ could be a class for the text to belong to ie. for sentiment analysis \textit{"positive"} and \textit{"neutral"} and thus the prediction of the classification model. Secondly, for translation, values $K$ and $V$ could be derived from the English sentence \textit{"Hello, how are you?"} and $Q$ the sequence "¿Hola, como estas?" for supervised English-Spanish machine translation. All of the State-of-the-Art models benchmarked in these experiments follow the concept of Multi-headed Attention. This is simply a concatenation of multiple $i$ attention heads $h_{i}$ to form a larger network of interconnected attention units:
\begin{equation}
\begin{aligned}
    MultiHead(Q,K,V) = Concatenate(head_{1}, ..., head_{h})W^{O} \\
    head_{i} = Attention(QW^{Q}_{i}, KW^{K}_{i}, VW^{V}_{i}).
\end{aligned}
\end{equation}

\begin{figure}
    \centering
    \includegraphics[scale=0.35]{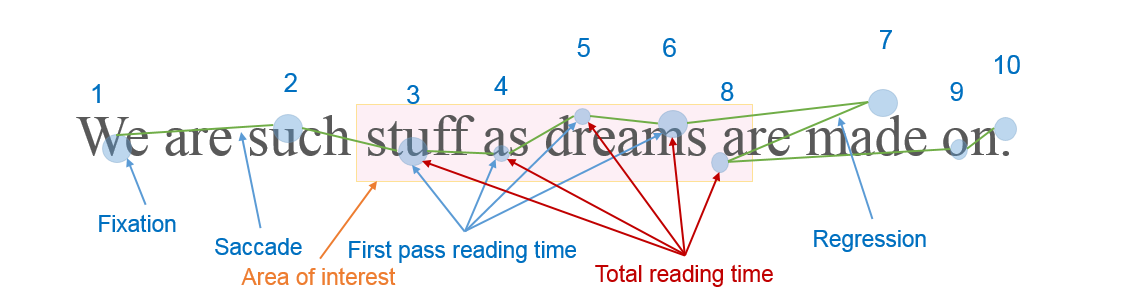}
    \caption{An eye-tracking study of natural reading from ~\cite{eckstein2019reading}. The reader's gaze naturally follows a left-to-right reading pattern with a fluctuation back to the main area of interest, where the main reading time is greater than that of the rest of the sentence.}
    \label{fig:reading-eye-tracker}
\end{figure}

It is important to note that human beings also do not read in a token-sequential nature as is with classical models such as the Long Short Term Memory (LSTM) network~\cite{hochreiter1997long}. Figure \ref{fig:reading-eye-tracker} from a 2019 study on reading comprehension~\cite{eckstein2019reading} shows human behaviour while reading. It can be observed from this example and other related studies~\cite{shagass1976eye,kruger2014subtitles,wang2019tracking}, that rather than simply reading left-to-right (or right-to-left ~\cite{wang2019tracking,marquis2020eyes}), instead attention is paid to areas of interest within the document. Of course, a human being does not follow the equations previously described, but it can be noted that attention-based models are more similar to human reading comprehension than that of sequential models such as the LSTM. Later, in Section \ref{sec:model-exploration}, during the exploration of top features within correct classifications, it can be observed that RoBERTa also focuses upon select areas of interest within a text for prediction. 

The Text-to-Text Transfer Transformer T5 model is a unified approach to text transformers from Google AI~\cite{raffel2019exploring}. T5 aims to unify NLP tasks by restricting output to text which is then interpreted to score the learning task; for example, it is natural to have a text output for a translation task (as per the previous example on English-Spanish translation), but for classification tasks on the other hand, a sparse vector for each prediction is often expected - T5 instead would output a textual representation of the class(es). This feature allows T5 to be extended to many NLP tasks outside of those suggested and benchmarked in the original work. To give a specific example to this study, an English-English translation of example \textit{"what time is it right now?"} to \textit{"could you tell me the time, please?"} provides a paraprhasing activity. That is, to express the same meaning of a text written in a different way. 

Chatbots are a method of human-machine interaction that have transcended novelty to become a useful technology of the modern world. A biological signal study from 2019 (Muscular activity, respiration, heart rate, and electrical behaviours of the skin) found that textual chatbots provide a more comfortable platform of interaction than with more human-like animated avatars, which caused participants to grow uncomfortable within the uncanny valley~\cite{ciechanowski2019shades}. Many chatbots exist as entertainment and as forms of art, such as in 2018~\cite{candello2018having} when natural interaction was enabled via state-of-art of the art methods for character generation from text~\cite{haller2013designing}. This allowed for 10,000 visitors to converse with 19th century characters from Machado de Assis' “Dom Casmurro”. It has been strongly suggested through multiple experiments that natural interaction with chatbots will provide a useful educational tool in the future for students of varying ages~\cite{kerlyl2006bringing,leonhardt2007using,leonhardt2007using,bollweg2018robots}. The main open issue in the field of conversational agents is data scarcity which in turn can lead to unrealistic and unnatural interaction, overcoming which are requirements for the Loebner Prize based on the Turing test~\cite{stephens2002has}. Solutions have been offered such as data selection of input~\cite{dimovski2018submodularity}, input simplification and generalisation~\cite{bird2018learning}, and more recently parapahrasing of data~\cite{virkar2019humanizing}. These recent advances in data augmentation by paraphrasing in particular have shown promise in improving conversational systems by increasing understanding of naturally spoken language~\cite{hou2018sequence,jin2018using}. 

\section{Proposed Approach}
\label{sec:method}
In this section, the proposed approach followed by the experiments are described, from data collection to modes of learning and classification.

The main aim of this work is to enable accessibility to previous studies, and in particular the machine learning models derived throughout them. Accessibility is presented in the form of social interaction, where a user requests to use a system in particular via natural language and the task is derived and performed. The seven commands are:
\begin{itemize}
    \item Scene Recognition ~\cite{bird2020look} - The participant requests a scene recognition algorithm to be instantiated, a camera and microphone are activated for multi-modality classification.
    \item EEG Classification - The participant requests an EEG classification algorithm to be instantiated and begins streaming data from a MUSE EEG headband, there are two algorithms:
    \begin{itemize}
        \item EEG Mental State Classification ~\cite{bird2018study} - Classification of whether the participant is concentrating, relaxed, or neutral.
        \item EEG Emotional State Classification ~\cite{bird2019mental} - Classification of emotional valence, positive, negative, or neutral.
    \end{itemize}
    \item Sentiment Analysis of Text ~\cite{bird2019high} - The participant requests the instantiation of a sentiment analysis classification algorithm for a given text.
    \item Sign Language Recognition ~\cite{bird2020british} - The participant requests to converse via sign language, a camera and Leap Motion and Leap Motion are activated for multi-modality classification. Sign language is now accepted as input to the task-classification layer of the chatbot.
    \item Conversational AI ~\cite{bird2018learning} - The participant requests to have a conversation, a chatbot program is executed.
    \item Joke Generator ~\cite{manurung2008construction,petrovic2013unsupervised} - The participant requests to hear a joke, a joke-generator algorithm is executed and output is printed. 
\end{itemize}
Each of the given commands are requested in the form of natural social interaction (either by keyboard input, speech converted to text, or sign language converted to text), and through accurate recognition, the correct algorithm is executed based on classification of the human input. Tasks such as sentiment analysis of text and emotional recognition of EEG brainwaves, and mental state recognition compared to emotional state recognition, are requested in similar ways and as such constitutes a difficult classification problem. For these problems, minute lingual details must be recognised in order to overcome ambiguity within informal communication.

\begin{figure}
    \centering
    \includegraphics[scale=1]{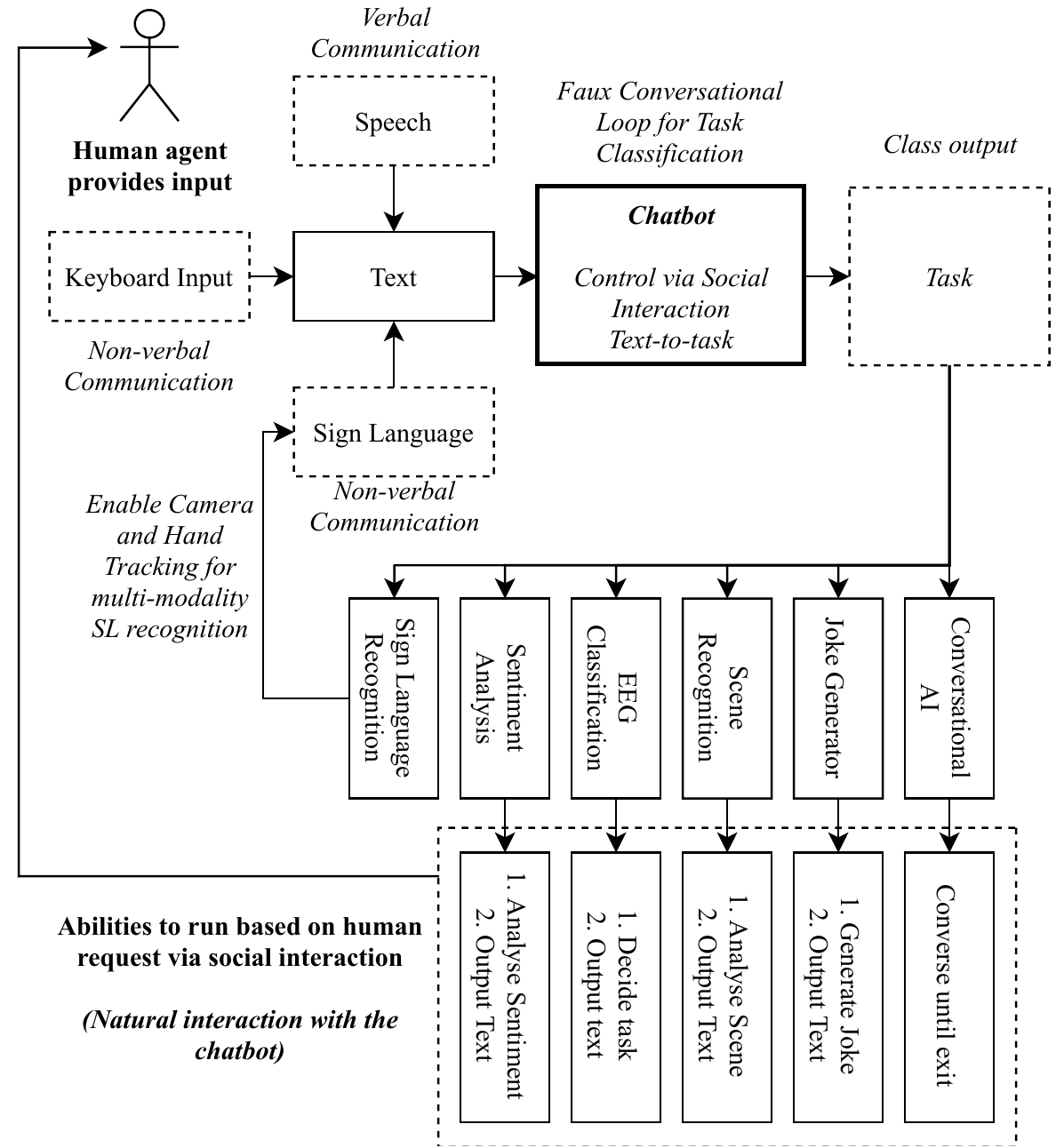}
    \caption{Overall view of the Chatbot Interaction with Artificial Intelligence (CI-AI) system as a looped process guided by human input, through natural social interaction due to the language transformer approach. The chatbot itself is trained via the process in Figure \ref{fig:training}.}
    \label{fig:chatbot}
\end{figure}

\begin{figure}
    \centering
    \includegraphics[scale=0.85]{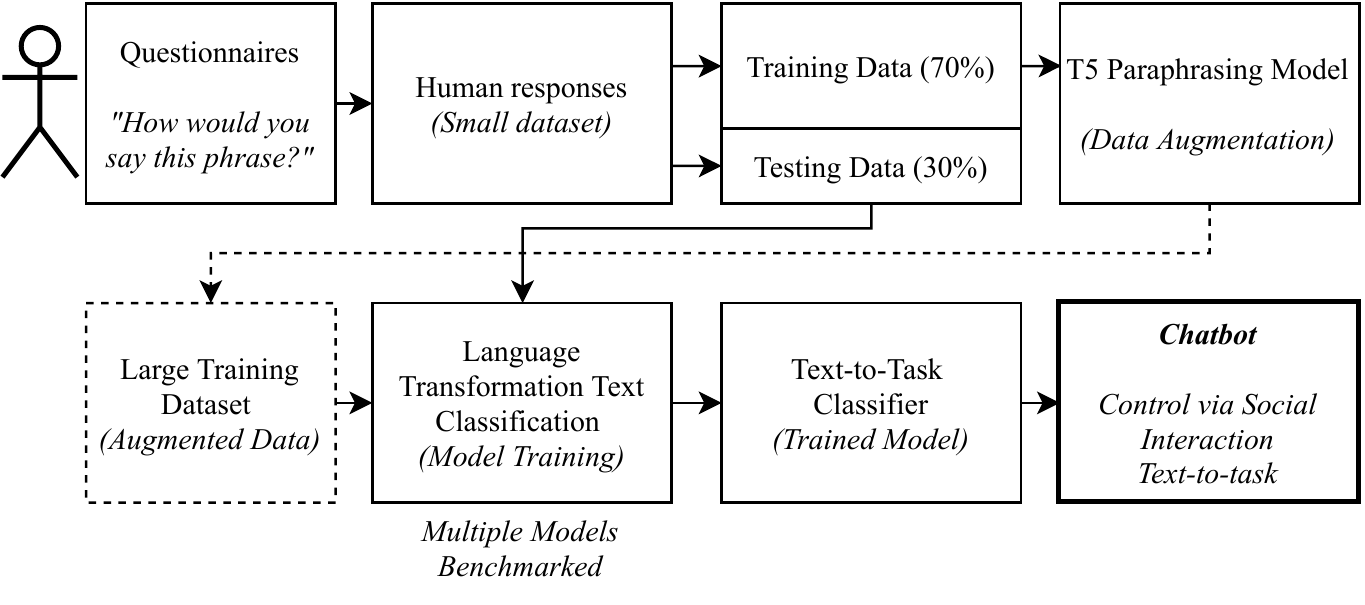}
    \caption{Data collection and model training process. In this example, the T5 paraphrasing model is used to augment and enhance the training dataset. Models are compared when they are augmented and when they are not on the same validation set, in order to discern what affect augmentation has.}
    \label{fig:training}
\end{figure}

Figure \ref{fig:chatbot} shows the overall view of the system. Keyboard input text, or speech and sign language converted to text provide an input of natural social interaction. The chatbot, trained on the tasks, classifies which task has been requested and executes said task for the human participant. Sign language, due to its need for an active camera and hand-tracking, is requested and activated via keyboard input or speech and itself constitutes a task. In order to derive the bold 'Chatbot' module in this figure, Figure \ref{fig:training} shows the training processes followed. Human data is gathered via questionnaires which gives a relatively small dataset (even though many responses were gathered, the nature of NLP tends to require a large amount of mined data), split into training and testing instances. The first experiment is built upon this data, and State-of-the-Art transformer classification models are benchmarked. In the second set of more complex experiments, the T5 paraphrasing model augments the training data and generates a large dataset, which are then also benchmarked with the same models and validation data in order to provide a direct comparison of the effects of augmentation. 

\begin{table}[]
\centering
\caption{A selection of example statements presented to the users for paraphrasing. One example is given for each for readability purposes, but a total of five examples were presented to the participants.}
\label{tab:questionnaire-examples}
\begin{tabular}{@{}ll@{}}
\toprule
\textbf{Example Statement}                             & \textbf{Class}       \\ \midrule
"Would you like to talk?"                              & CHAT                 \\
"Tell me a joke"                                       & JOKE                 \\
"Can you tell what mood I'm in from my brainwaves?"    & EEG-EMOTIONS         \\
"Am I concentrating? Or am I relaxed?                  & EEG-MENTAL-STATE     \\
"Look around and tell me where you are."               & SCENE-CLASSIFICATION \\
"Is this message being sarcastic or are they genuine?" & SENTIMENT-ANALYSIS   \\
"I cannot hear the audio, please sign instead."        & SIGN-LANGUAGE        \\ \bottomrule
\end{tabular}
\end{table}

A questionnaire was published online for users to provide human data in the form of examples of commands that would lead to a given task classification. Five examples were given for each, and Table \ref{tab:questionnaire-examples} shows some examples that were presented. The questionnaire instructions were introduced with \textit{"For each of these questions, please write how you would state the text differently to how the example is given. That is, paraphrase it. Please give only one answer for each. You can be as creative as you want!"}. Two examples were given that were not part of any gathered classes, \textit{"If the question was: 'How are you getting to the cinema?' You could answer: 'Are we driving to the cinema or are we getting the bus?'} and \textit{"If the question was: 'What time is it?' You could answer: 'Oh no, I slept in too late... Is it the morning or afternoon? What's the time?'"}. These examples were designed to show the users that creativity and diversion from the given example was not just acceptable but also encouraged, so long as the general meaning and instruction of and within the message was retained (the instructions ended with \textit{"The example you give must still make sense, leading to the same outcome."}). Extra instructions were given as and when requested, and participants did not submit any example phrases nor were any duplicates submitted. A total of 483 individual responses were recorded. The answers were split 70/30 on a per-class basis to provide two class-balanced datasets, firstly for training (and augmentation), and secondly for validation. That is, regardless of augmentation, the model is tested based on this validation set and are all thus directly comparable in terms of their learning abilities. 

The T5 paraphrasing model which was trained on the Quora question pairs dataset~\cite{QuoraQue87:online} is executed a maximum of 50 times for each statement within the training set, where the model will stop generating paraphrases if the limit of possibilities or 50 total are reached. Once each statement had been paraphrased, a random subsample of the dataset on a per-class basis was taken set at the number of data objects within the least common class (sign language). Concatenated then with the real training data, a dataset of 13,090 examples were formed (1870 per class). This dataset thus constitutes the second training set for the second experiment, in order to compare the effects of data augmentation for the problem presented.

\begin{table}[]
\centering
\caption{An overview of models benchmarked and their topologies}
\label{tab:topologies}
\begin{tabular}{@{}ll@{}}
\toprule
\textbf{Model}                  & \textbf{Topology}                                                                                            \\ \midrule
\textit{\textbf{BERT}}~\cite{devlin2018bert}          & 12-layer, 768-hidden, 12-heads, 110M parameters.                                                             \\
\textit{\textbf{DistilBERT}}~\cite{sanh2019distilbert}    & 6-layer, 768-hidden, 2-heads, 66M parameters                                                                 \\
\textit{\textbf{RoBERTa}}~\cite{liu2019roberta}       & 12-layer, 768-hidden, 12-heads, 125M parameters                                                              \\
\textit{\textbf{DistilRoBERTa}}~\cite{liu2019roberta,Wolf2019HuggingFacesTS} & 6-layer, 768-hidden, 12-heads, 82M parameters                                                                \\
\textit{\textbf{XLM}}~\cite{conneau2019cross}           & 12-layer, 2048-hidden, 16-heads, 342M parameters                                                                           \\
\textit{\textbf{XLM-RoBERTa}}~\cite{conneau2019unsupervised}   & \begin{tabular}[c]{@{}l@{}}12-layer, 768-hidden, 3072 feed-forward, \\ 8-heads, 125M parameters\end{tabular} \\
\textit{\textbf{XLNet}}~\cite{yang2019xlnet}         & \begin{tabular}[c]{@{}l@{}}12-layer, 768-hidden, \\ 12-heads, 110M parameters\end{tabular}                   \\ \bottomrule
\end{tabular}
\end{table}

Table \ref{tab:topologies} shows the models that are trained and benchmarked on the two training sets (Human, Human+T5), and validated on the same validation dataset. It can be observed that the models are complex, and training requires a relatively high amount of computational resources. Due to this, the pre-trained weights for each model are fine-tuned for two epochs on each of the training datasets. 

\subsection{Statistical Ensemble of Transformer Classifiers}
\begin{figure}
    \centering
    \includegraphics[scale=1]{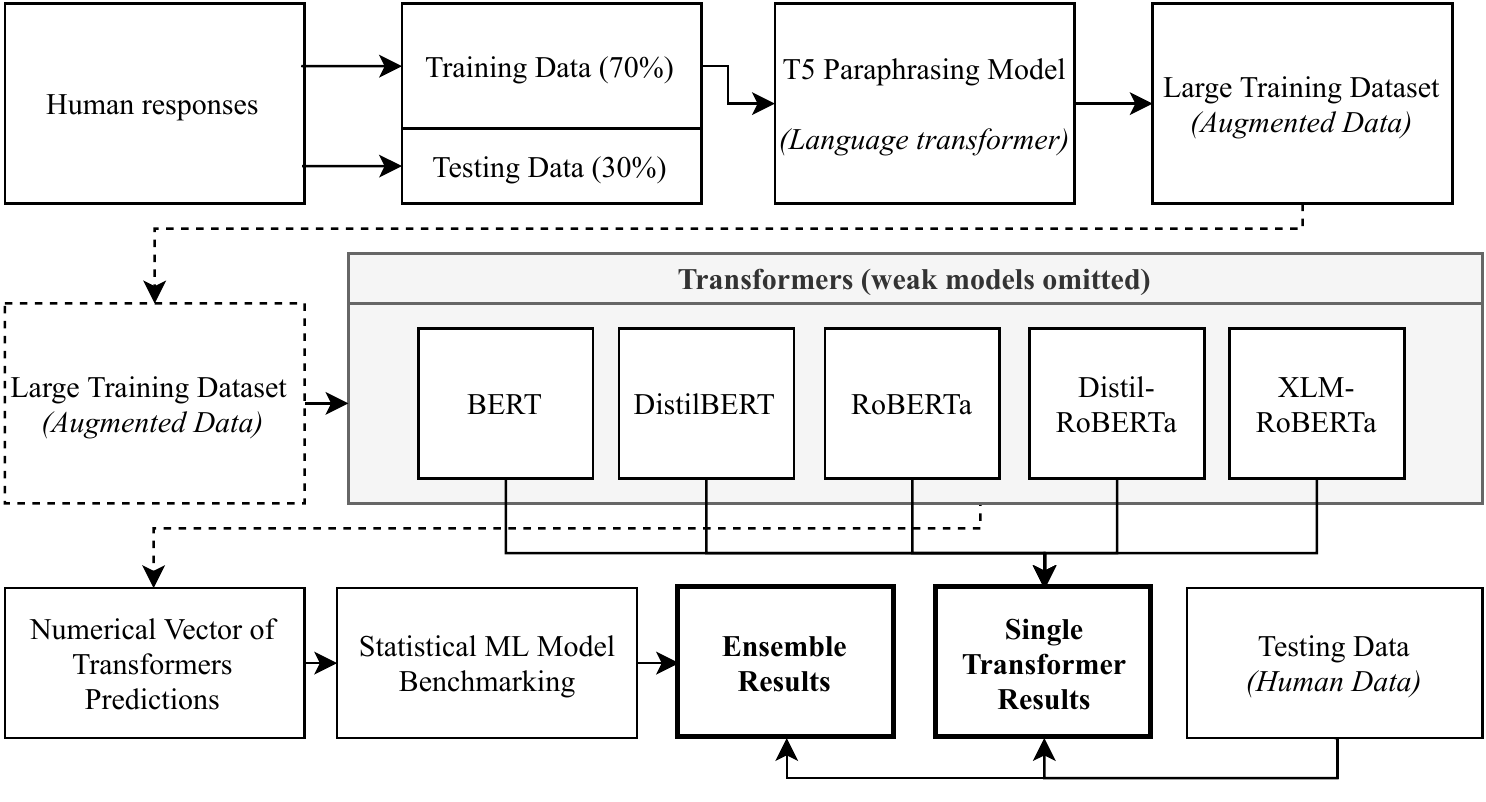}
    \caption{An ensemble strategy where statistical machine learning models trained on the predictions of the transformers then classify the text based on the test data predictions of the transformer classification models.}
    \label{fig:ensemble-diagram}
\end{figure}
Following the results detailed later in Section \ref{sec:results}, two main findings were made; 1) that all models were improved by T5 augmentation and 2) XLM and XLNet were weak solutions to the problem and scored relatively low classification scores. Following these findings, an extension to the study through an ensemble method is devised which combines the five strong models when trained on paraphrased data, which can be observed in Figure \ref{fig:ensemble-diagram}. The training and test datasets are firstly distilled into a numerical vector of five predictions made by the five transformer models. Following this, statistical machine learning models are trained on the training set and validated by the test set in order to discern whether combining the models together ultimately improves the ability of the model. The reasoning behind a statistical ensemble is that it enables possible improvements to a decision system’s robustness and accuracy~\cite{zhang2012ensemble}. Given that nuanced differences between the transformers may lead to 'personal' improvements in some situations and negative impacts in others, for example when certain phrases appear within commands, a more democratic approach may allow the pros of some models outweigh the cons of others. Employing a statistical model to learn these patterns by classifying the class based on the outputs of the previous models would thus allow said ML model to learn these nuanced differences between the transformers.

\section{Experimental Hardware and Software}
The experiments were executed on an NVidia Tesla K80 GPU which has 4992 CUDA cores and 24 GB of GDDR5 memory via the Google Colab platform. The Transformers were implemented via the KTrain library~\cite{maiya2020ktrain}, which is a back-end for TensorFlow~\cite{tensorflow2015-whitepaper} Keras~\cite{chollet2015keras}. The pretrained weights for the Transformers prior to fine-tuning were from the HuggingFace NLP Library~\cite{Wolf2019HuggingFacesTS}.

The pre-trained T5 paraphrasing model weights were from ~\cite{EllaChan53:online}. The model was implemeted with the HuggingFace NLP Library~\cite{Wolf2019HuggingFacesTS} via PyTorch~\cite{NEURIPS2019_9015} and was trained for two epochs ($\sim$20 hours) on the p2.xlarge AWS ec2.

The statistical models for the ensemble results were implemented in Python via the Scikit-learn toolkit~\cite{scikit-learn} and executed on an Intel Core i7 Processor (3.7GHz). 

\section{Results}
\label{sec:results}

\begin{table}[]
\centering
\caption{Classification results of each model on the same validation set, both with and without augmented paraphrased data within the training dataset. Bold shows best model per run, underline shows the best model overall.}
\label{tab:results}
\begin{tabular}{@{}lllllllll@{}}
\toprule
\multirow{2}{*}{\textbf{Model}} & \multicolumn{4}{l}{\textbf{With T5 Paraphrasing}}                                                & \multicolumn{4}{l}{\textbf{Without T5 Paraphrasing}}                                             \\ \cmidrule(l){2-9} 
                                & \textit{\textbf{Acc.}} & \textit{\textbf{Prec.}} & \textit{\textbf{Rec.}} & \textit{\textbf{F1}} & \textit{\textbf{Acc.}} & \textit{\textbf{Prec.}} & \textit{\textbf{Rec.}} & \textit{\textbf{F1}} \\ \cmidrule(r){1-1}
\textit{\textbf{BERT}}          & \textbf{98.55}         & 0.99                    & 0.99                   & 0.99                 & 90.25                  & 0.93                    & 0.9                    & 0.9                  \\
\textit{\textbf{DistilBERT}}    & \textbf{98.34}         & 0.98                    & 0.98                   & 0.98                 & 97.3                   & 0.97                    & 0.97                   & 0.97                 \\
\textit{\textbf{DistilRoBERTa}} & \textbf{98.55}         & 0.99                    & 0.99                   & 0.99                 & 95.44                  & 0.96                    & 0.95                   & 0.95                 \\
\textit{\textbf{RoBERTa}}       & {\ul \textbf{98.96}}   & 0.99                    & 0.99                   & 0.99                 & 97.93                  & 0.98                    & 0.98                   & 0.98                 \\
\textit{\textbf{XLM}}           & \textbf{14.81}         & 0.15                    & 0.15                   & 0.15                 & 13.69                  & 0.02                    & 0.14                   & 0.03                 \\
\textit{\textbf{XLM-RoBERTa}}   & \textbf{98.76}         & 0.99                    & 0.99                   & 0.99                 & 87.97                  & 0.9                     & 0.88                   & 0.88                 \\
\textit{\textbf{XLNet}}         & \textbf{35.68}         & 0.36                    & 0.35                   & 0.36                 & 32.99                  & 0.33                    & 0.24                   & 0.24                 \\ \midrule
\textit{\textbf{Average}}       & 77.66                  & 0.78                    & 0.78                   & 0.78                 & 73.65                  & 0.73                    & 0.72                   & 0.71                 \\ \bottomrule
\end{tabular}
\end{table}

\begin{table}[]
\centering
\caption{Observed increases in training metrics for each model due to data augmentation via paraphrasing the training dataset.}
\label{tab:results-change}
\begin{tabular}{@{}lllll@{}}
\toprule
\multirow{2}{*}{\textbf{Model}} & \multicolumn{4}{l}{\textbf{Increase of Metrics}}                          \\ \cmidrule(l){2-5} 
                                & \textit{\textbf{Acc.}} & \textit{\textbf{Prec.}} & \textit{\textbf{Rec.}} & \textit{\textbf{F1}} \\ \cmidrule(r){1-1}
\textit{\textbf{BERT}}          & 8.3                    & 0.06                    & 0.09                   & 0.09                 \\
\textit{\textbf{DistilBERT}}    & 1.04                   & 0.01                    & 0.01                   & 0.01                 \\
\textit{\textbf{DistilRoBERTa}} & 3.11                   & 0.03                    & 0.04                   & 0.04                 \\
\textit{\textbf{RoBERTa}}       & 1.03                   & 0.01                    & 0.01                   & 0.01                 \\
\textit{\textbf{XLM}}           & 1.12                   & 0.13                    & 0.01                   & 0.12                 \\
\textit{\textbf{XLM-RoBERTa}}   & 10.79                  & 0.09                    & 0.11                   & 0.11                 \\
\textit{\textbf{XLNet}}         & 2.69                   & 0.03                    & 0.11                   & 0.12                 \\ \midrule
\textit{\textbf{Average}}       & 4.01                   & 0.05                    & 0.05                   & 0.07                 \\ \bottomrule
\end{tabular}
\end{table}

Table \ref{tab:results} shows the overall results for all of the experiments. Every single model, even the weakest XLNet for this particular problem, was improved when training on the human data alongside the augmented data which can be seen for the increases in metrics in Table \ref{tab:results-change}. This required a longer training time due to the more computationally intense nature of training on a larger dataset. T5 paraphrasing for data augmentation led to an average accuracy increase of 4.01 points, and the precision, recall, and F1 scores were also improved at an average of 0.05, 0.05, and 0.07, respectively.

The best performing model was RoBERTa when training on the human training set as well as the augmented data. This model achieved 98.96\% accuracy with 0.99 precision, recall and F1 score. The alternative to training only on the human data achieved 97.93\% accuracy with stable precision, recall and F1 scores of 0.98. The second best performing models were both the distilled version of RoBERTa and BERT, which achieved 98.55\% and likewise 0.98 for the other three metrics. Interestingly, some models saw a drastic increase in classification ability when data augmentation was implemented; the BERT model rose from 90.25\% classification accuracy with 0.93 precision, 0.9 recall and 0.9 F1 score with a +8.3\% increase and then more stable metrics of 0.99 each as described previously. In the remainder of this section, the 98.96\% performing RoBERTa model when trained upon human and T5 data is explored further. This includes, exploration of errors made overall and per specific examples, as well as an exploration of top features within successful predictions made. 

\begin{figure}
    \centering
    \includegraphics[scale=0.9]{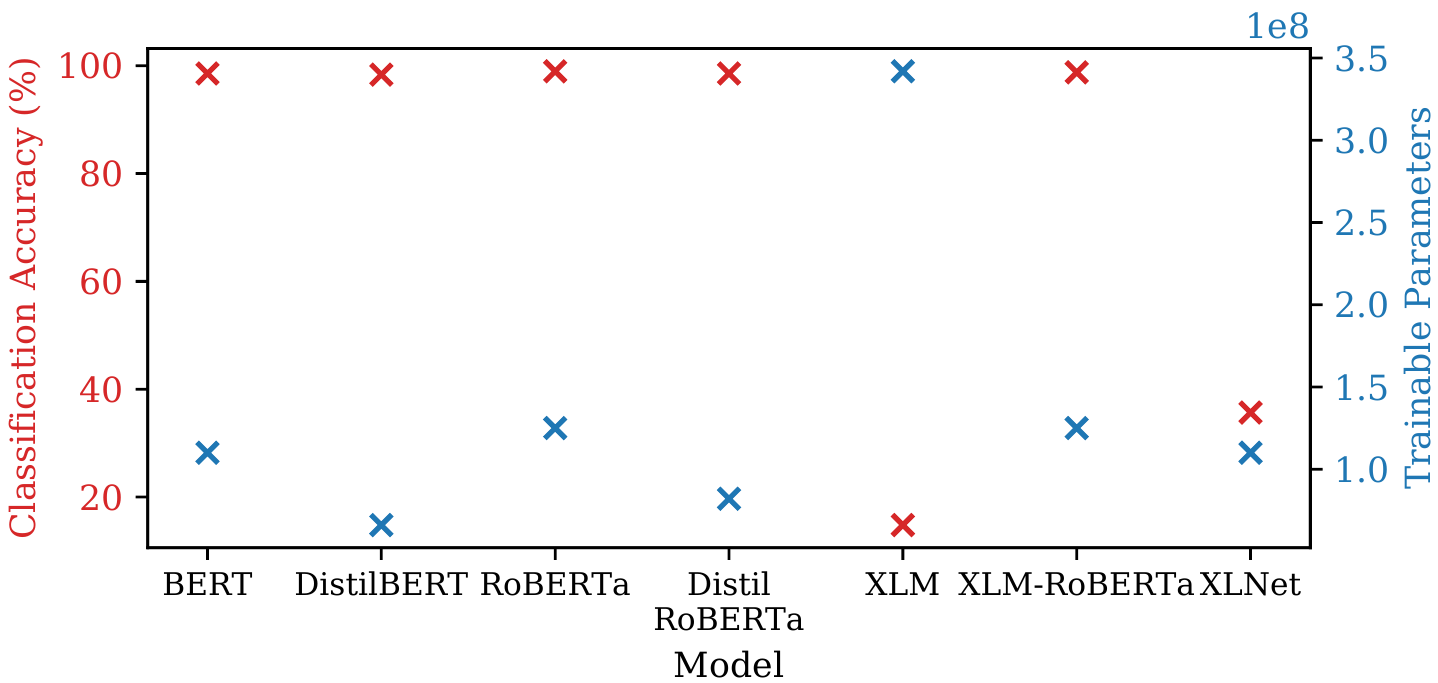}
    \caption{Comparison of each model's classification ability and number of million trainable parameters within them.}
    \label{fig:accuracy-vs-params}
\end{figure}

Figure \ref{fig:accuracy-vs-params} shows a comparison between the model performance and number of trainable parameters. Note that the most complex model scored the least in terms of classification ability. The best performing model was the third most complex model of all. The least complex model, DistilBERT, achieved a relatively high accuracy of 98.34\%. 

\section{Exploration of the best transformer model}
\label{sec:model-exploration}
In this section, we explore the best model. The best model, as previously discussed, was the RoBERTa model when training on both the collected training data and the paraphrased data generated by the T5 model. 

\begin{table}[]
\centering
\caption{Per-class precision, recall, and F1 score metrics for the best model.}
\label{tab:per-class}
\begin{tabular}{@{}llll@{}}
\toprule
\textbf{Class}                         & \textbf{Prec.} & \textbf{Rec.} & \textbf{F1} \\ \midrule
\textit{\textbf{CHAT}}                 & 1.00           & 0.99          & 0.99        \\
\textit{\textbf{EEG-EMOTIONS}}         & 0.99           & 0.97          & 0.98        \\
\textit{\textbf{EEG-MENTAL-STATE}}     & 0.99           & 1.00          & 0.99        \\
\textit{\textbf{JOKE}}                 & 0.98           & 0.98          & 0.98        \\
\textit{\textbf{SCENE-CLASSIFICATION}} & 1.00           & 1.00          & 1.00        \\
\textit{\textbf{SENTIMENT-ANALYSIS}}   & 0.97           & 0.99          & 0.98        \\
\textit{\textbf{SIGN-LANGUAGE}}        & 1.00           & 1.00          & 1.00        \\ \bottomrule
\end{tabular}
\end{table}

\begin{figure}
    \centering
    \includegraphics[scale=0.9]{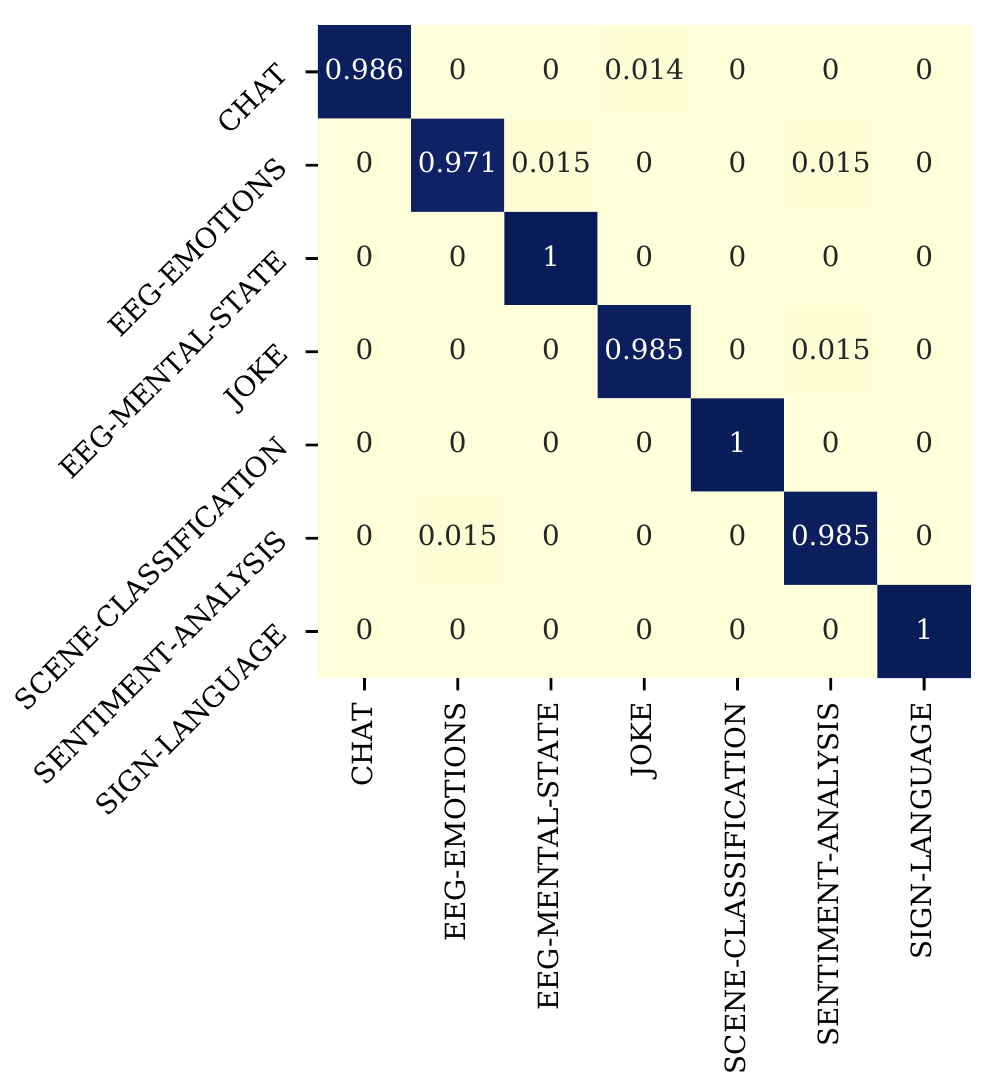}
    \caption{Normalised confusion matrix for the best command classification model, which was RoBERTa when trained on human data and augmented T5 paraphrased data.}
    \label{fig:roberta-confusion}
\end{figure}

Table \ref{tab:per-class} shows the classification metrics for each individual class by the RoBERTa model. The error matrix for the validation data can be seen in Figure \ref{fig:roberta-confusion}. The tasks of EEG mental state classification, scene recognition, and sentiment analysis were classified perfectly. Of the imperfect classes, the task of conversational AI ('CHAT') was sometimes misclassified as a request for a joke, which is likely due to the social nature of the two activities. EEG emotional state classification was rarely mistakenly classified as the mental state recognition and sentiment analysis tasks, firstly due to the closely related EEG tasks and secondly as sentiment analysis since data often involved terms synonymous with valence or emotion. Similarly, the joke class was also rarely misclassified as sentiment analysis, for example, "tell me something funny" and "can you read this email and tell me if they are being funny with me?" ('funny' in the second context being a British slang term for sarcasm). The final class with misclassified instances was sentiment analysis, as emotional state recognition, for the same reason previously described when the error occurred vice-versa. 

\subsection{Mistakes and probabilities}
In this section, we explore the biggest errors made when classifying the validation set by considering their losses.

\begin{table}[]
\centering
\caption{The most confusing sentences according to the model (all of those with a loss \textgreater 1) and the probabilities as to which class they were predicted to belong to. Key - C1: CHAT, C2: EEG-EMOTIONS, C3: EEG-MENTAL-STATE, C4: JOKE, C5: SCENE-RECOGNITION, C6: SENTIMENT-ANALYSIS, C7: SIGN-LANGUAGE}
\label{tab:mistakes}
\begin{tabular}{@{}llllllll@{}}
\toprule
\textbf{Text}                     & \multicolumn{7}{l}{\textit{"What is your favourite one liner?"}}                                                                                                                  \\ \midrule
\textbf{Actual}                     & \multicolumn{7}{l}{C4}                                                                                                                                                          \\
\textbf{Predicted}                & \multicolumn{7}{l}{C6}                                                                                                                                            \\
\textbf{Loss}                     & \multicolumn{7}{l}{6.24}                                                                                                                                                          \\
\multirow{2}{*}{\textbf{Prediction Probabilities}} & \textit{\textbf{C1}} & \textit{\textbf{C2}} & \textit{\textbf{C3}} & \textit{\textbf{C4}} & \textit{\textbf{C5}} & \textit{\textbf{C6}} & \textit{\textbf{C7}} \\
                                  & 0.0163               & 0.001                & 0                 & 0.002               & 0.001                & 0.977                & 0.002               \\ \midrule
\textbf{Text}                     & \multicolumn{7}{l}{\textit{"What is your favourite movie?"}}                                                                                                                      \\ \midrule
\textbf{Actual}                     & \multicolumn{7}{l}{C1}                                                                                                                                                          \\
\textbf{Predicted}                & \multicolumn{7}{l}{C4}                                                                                                                                                          \\
\textbf{Loss}                     & \multicolumn{7}{l}{2.75}                                                                                                                                                          \\
\multirow{2}{*}{\textbf{Prediction Probabilities}} & \textit{\textbf{C1}} & \textit{\textbf{C2}} & \textit{\textbf{C3}} & \textit{\textbf{C4}} & \textit{\textbf{C5}} & \textit{\textbf{C6}} & \textit{\textbf{C7}} \\
                                  & 0.064                  & 0.0368                  & 0.007                    & 0.513                  & 0.338                   & 0.022                   & 0.02                   \\ \midrule
\textbf{Text}                     & \multicolumn{7}{l}{\textit{"How do I feel right now?"}}                                                                                                                           \\ \midrule
\textbf{Actual}                     & \multicolumn{7}{l}{C1}                                                                                                                                                          \\
\textbf{Predicted}                & \multicolumn{7}{l}{C4}                                                                                                                                                          \\
\textbf{Loss}                     & \multicolumn{7}{l}{2.75}                                                                                                                                                          \\
\multirow{2}{*}{\textbf{Prediction Probabilities}} & \textit{\textbf{C1}} & \textit{\textbf{C2}} & \textit{\textbf{C3}} & \textit{\textbf{C4}} & \textit{\textbf{C5}} & \textit{\textbf{C6}} & \textit{\textbf{C7}} \\
                                  & 0.007                  & 0.01                   & 0.352                    & 0.434                   & 0.016                   & 0.176                   & 0.005                  \\ \midrule
\textbf{Text}                     & \multicolumn{7}{l}{\textit{"Run emotion classification"}}                                                                                                                         \\ \midrule
\textbf{Actual}                     & \multicolumn{7}{l}{C6}                                                                                                                                            \\
\textbf{Predicted}                & \multicolumn{7}{l}{C2}                                                                                                                                                  \\
\textbf{Loss}                     & \multicolumn{7}{l}{1.71}                                                                                                                                                          \\
\multirow{2}{*}{\textbf{Prediction Probabilities}} & \textit{\textbf{C1}} & \textit{\textbf{C2}} & \textit{\textbf{C3}} & \textit{\textbf{C4}} & \textit{\textbf{C5}} & \textit{\textbf{C6}} & \textit{\textbf{C7}} \\
                                  & 0                      & 0.672                    & 0.001                    & 0.002                  & 0.004                   & 0.32                   & 0                      \\ \midrule
\textbf{Text}                     & \multicolumn{7}{l}{\textit{"What is the valence of my brainwaves?"}}                                                                                                              \\ \midrule
\textbf{Actual}                     & \multicolumn{7}{l}{C2}                                                                                                                                                  \\
\textbf{Predicted}                & \multicolumn{7}{l}{C3}                                                                                                                                              \\
\textbf{Loss}                     & \multicolumn{7}{l}{1.05}                                                                                                                                                          \\
\multirow{2}{*}{\textbf{Prediction Probabilities}} & \textit{\textbf{C1}} & \textit{\textbf{C2}} & \textit{\textbf{C3}} & \textit{\textbf{C4}} & \textit{\textbf{C5}} & \textit{\textbf{C6}} & \textit{\textbf{C7}} \\
                                  & 0.001               & 0.349                & 0.647                 & 0.001               & 0.001                & 0.002                & 0               \\ \bottomrule
\end{tabular}
\end{table}

\begin{figure}
    \centering
    \includegraphics[scale=1]{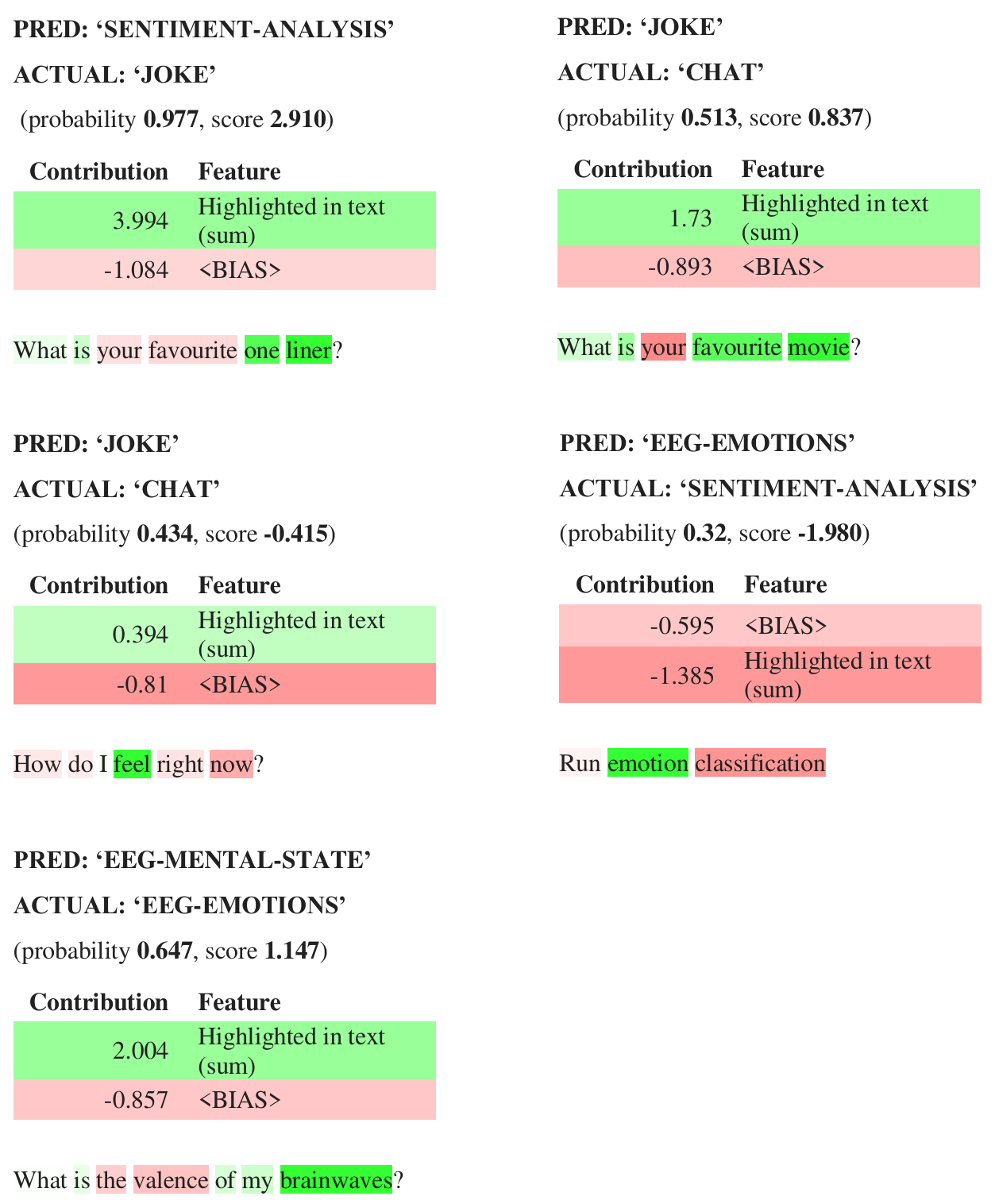}
    \caption{Exploration and explanation for the errors made during validation which had a loss \textgreater 1 (five such cases).}
    \label{fig:error-exploration}
\end{figure}

Table \ref{tab:mistakes} shows the most confusing data objects within the training set and Figure \ref{fig:error-exploration} explores which parts of the phrase the model focused on to derive these erroneous classifications. Overall, only five misclassified sentences had a loss above 1; the worst losses were in the range of 1.05 to 6.24. The first phrase, "what is your favourite one liner?", may likely have caused confusion due to the term "one liner" which was not present within the training set. Likewise, the term "valence" in "What is the valence of my brainwaves?" was also not present within the training set, and the term "brainwaves" was most common when referring to mental state recognition rather than emotional state recognition. 

An interesting error occurred from the command "Run emotion classification", where the classification was incorrectly given as EEG emotional state recognition rather than Sentiment Analysis. The command collected from a human subject was ambiguous, and as such the two most likely classes were the incorrect EEG Emotions at a probability of 0.672 and the correct Sentiment Analysis at a probability of 0.32. This raises an issue to be explored in future works, given the nature of natural social interaction, it is likely that ambiguity will be present during conversation. Within this erroneous classification, two classes were far more likely than all other classes present, and thus a choice between the two in the form of a question akin to human deduction of ambiguous language would likely solve such problems and increase accuracy. Additionally, this would rarely incur the requirement of further effort from the user. 

\subsection{Top features within unseen data}
Following the training of the model, this section explores features within data when an unseen phrase or command is uttered. That is, the examples given in this section were not data within the training or validation datasets, and thus are more accurate simulations of the model within a real-world scenario given new data to process based on the rules learnt during training. 
\begin{figure}
    \centering
    \includegraphics[scale=1]{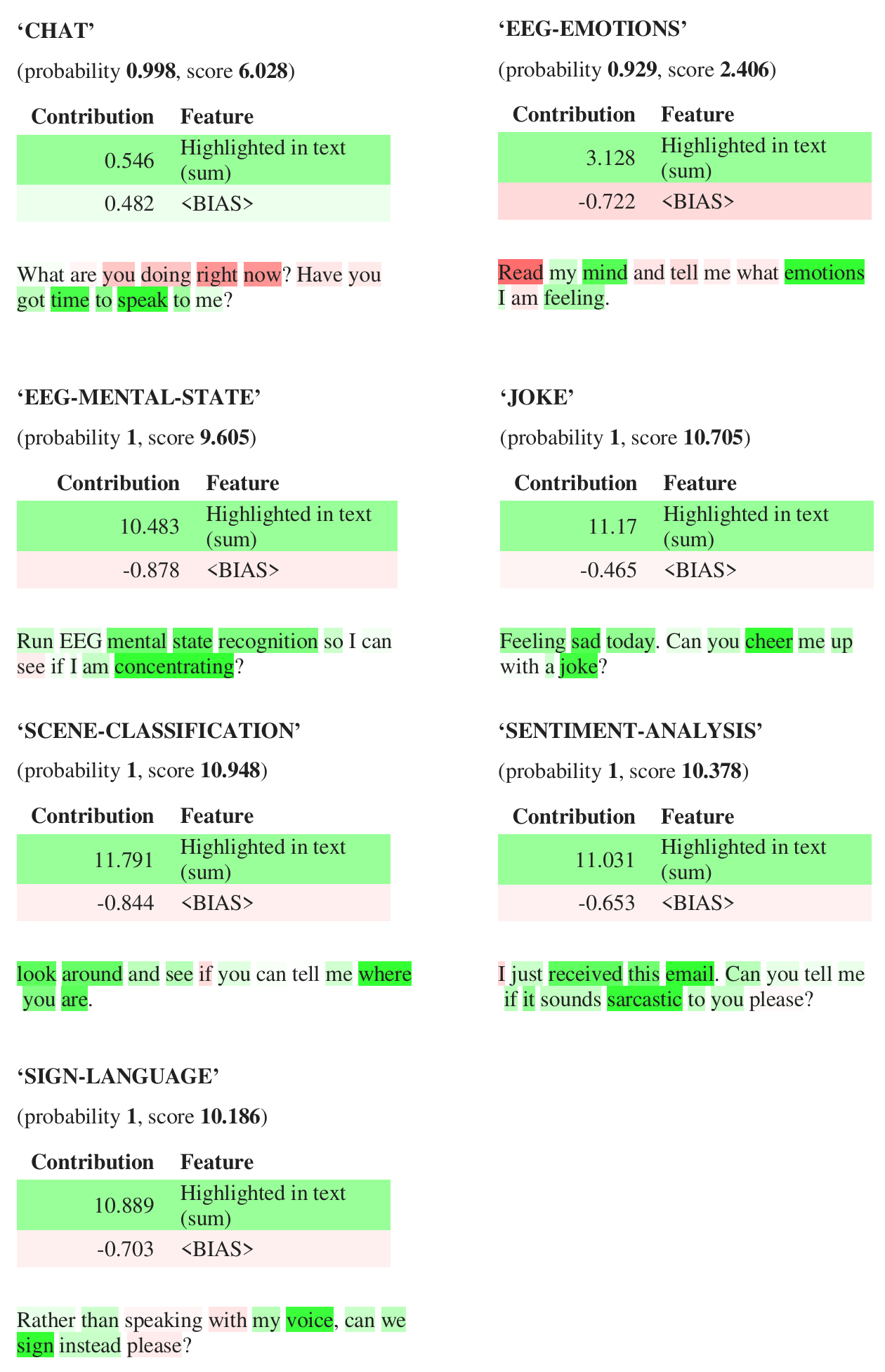}
    \caption{Exploration of the best performing model by presenting unseen sentences and explaining predictions. Green denotes useful features and red denotes features useful for another class (detrimental to probability).}
    \label{fig:exploring-unseen}
\end{figure}

In this regard, Figure \ref{fig:exploring-unseen} shows an example of a correct prediction of an unseen data's class, for each class. Interestingly, the model shows behaviour reminiscent of human reading~\cite{biedert2012robust,kunze2013my} due to transformers not being limited to considering a temporal sequence in chronological order of appearance. 

In the first example the most useful features were 'time to speak' followed by 'got', 'to' and 'me'. The least useful features were 'right now', which alone would be classified as 'SCENE-CLASSIFICATION' with a probability of 0.781 due to many provided training examples for such class containing questions such as 'where are you \textbf{right now}? Can you run scene recognition and tell me?'. The second example also had a strong negative impact from the word 'read' which alone would be classified as 'SENTIMENT-ANALYSIS' with a probability of 0.991 due to the existence of phrases such as 'please \textbf{read} this message \textbf{and tell me} if they are angry with me' being popular within the gathered human responses and as such the augmented data. This example found correct classification due to the terms 'emotions' and 'mind' primarily, followed by 'feeling'. Following these two first examples, the remaining five examples were strongly classified. In the mental state recognition task, even though the term 'mental state' was specifically uttered, the term 'concentrating' was the strongest feature within the statement given the goal of the algorithm to classify concentrating and relaxed states of mind. As could be expected, the 'JOKE' task was best classified by the term 'joke' itself being present, but, interestingly, the confidence of classification was increased with the phrases 'Feeling sad today.' and 'cheer me up'. The scene classification task was confidently predicted with a probability of 1 mainly due to the terms 'look around' and 'where you are'. The red highlight for the word 'if' alone would be classified as 'SENTIMENT-ANALYSIS' with a probability of 0.518 given the popularity of phrases along the lines of '\textbf{if} they are \textit{emotion} or \textit{emotion}'. 

The sentiment analysis task was then, again, confidently classified correctly with a probability of 1. This was due to the terms 'received this email', 'if', and 'sarcastic' being present. Finally, the sign language task was also classified with a probability of 1 most due to the features 'voice' and 'sign'. The red features highlighted, 'speaking with please' would alone be classified as 'CHAT' with a probability of 0.956, since they are strongly reminiscent to commands such as, 'can we speak about something please?'. 

An interesting behaviour to note from these examples is the previously described nature of reading. Transformer models are advancing the field of NLP in part thanks due to their lack of temporal restriction, ergo the limitations existent within models such as Recurrent or Long Short Term Memory Neural Networks. This allows for behaviours more similar to a human being, such as when someone may focus on certain key words first before glancing backwards for more context. Such behaviours are not possible with sequence-based text classification techniques.

\subsection{Transformer Ensemble Results}
\begin{table}[]
\centering
\caption{Information Gain ranking of each predictor model by 10 fold cross validation on the training set}
\label{tab:ensemble-infogain}
\begin{tabular}{@{}lll@{}}
\toprule
\textbf{\begin{tabular}[c]{@{}l@{}}Predictor Model\\ (Transformer)\end{tabular}} & \textbf{Average Ranking} & \textbf{Information Gain of Predictions} \\ \midrule
\textit{\textbf{BERT}}                                                           & 1 ($\pm$ 0)                    & 2.717 ($\pm$ 0.002)                            \\
\textit{\textbf{DistilBERT}}                                                     & 2 ($\pm$ 0)                    & 2.707 ($\pm$ 0.002)                            \\
\textit{\textbf{DistilRoBERTa}}                                                  & 3.1 ($\pm$ 0.3)                & 2.681 ($\pm$ 0.001)                            \\
\textit{\textbf{RoBERTa}}                                                        & 3.9 ($\pm$ 0.3)                & 2.676 ($\pm$ 0.003)                            \\
\textit{\textbf{XLM-RoBERTa}}                                                    & 5 ($\pm$ 0)                    & 2.653 ($\pm$ 0.002)                            \\ \bottomrule
\end{tabular}
\end{table}

\begin{table}[]
\centering
\caption{Results for the ensemble learning of Transformer predictions compared to the best single model (RoBERTa)}
\label{tab:ensemble-results}
\begin{tabular}{@{}llllll@{}}
\toprule
\textbf{Ensemble Method}                                                                     & \textbf{Accuracy} & \textbf{Precision} & \textbf{Recall} & \textbf{F1}    & \textbf{\begin{tabular}[c]{@{}l@{}}Difference over \\ RoBERTa\end{tabular}} \\ \midrule
\textit{\textbf{\begin{tabular}[c]{@{}l@{}}Logistic \\ Regression\end{tabular}}}             & \textbf{99.59}    & \textbf{0.996}     & \textbf{0.996}  & \textbf{0.996} & \textbf{+0.63}                                                              \\
\textit{\textbf{\begin{tabular}[c]{@{}l@{}}Random \\ Forest\end{tabular}}}                   & \textbf{99.59}    & \textbf{0.996}     & \textbf{0.996}  & \textbf{0.996} & \textbf{+0.63}                                                              \\
\textit{\textbf{\begin{tabular}[c]{@{}l@{}}Multinomial \\ Na\"{i}ve Bayes\end{tabular}}}         & 99.38             & 0.994              & 0.994           & 0.994          & +0.42                                                                       \\
\textit{\textbf{\begin{tabular}[c]{@{}l@{}}Bernoulli \\ Na\"{i}ve Bayes\end{tabular}}}           & 99.38             & 0.994              & 0.994           & 0.994          & +0.42                                                                       \\
\textit{\textbf{\begin{tabular}[c]{@{}l@{}}Linear \\ Discriminant \\ Analysis\end{tabular}}} & 99.38             & 0.994              & 0.994           & 0.994          & +0.42                                                                       \\
\textit{\textbf{XGBoost}}                                                                    & 99.38             & 0.994              & 0.994           & 0.994          & +0.42                                                                       \\
\textit{\textbf{\begin{tabular}[c]{@{}l@{}}Support \\ Vector Classifier\end{tabular}}}       & 99.38             & 0.994              & 0.994           & 0.994          & +0.42                                                                       \\
\textit{\textbf{\begin{tabular}[c]{@{}l@{}}Bayesian \\ Network\end{tabular}}}                & 99.38             & 0.994              & 0.994           & 0.994          & +0.42                                                                       \\
\textit{\textbf{\begin{tabular}[c]{@{}l@{}}Gaussian \\ Na\"{i}ve Bayes\end{tabular}}}            & 98.55             & 0.986              & 0.985           & 0.986          & -0.41                                                                       \\ \bottomrule
\end{tabular}
\end{table}

Following the previous findings, the five strongest models which were BERT (98.55\%), DistilBERT (98.34\%), RoBERTa (98.96\%), Distil-RoBERTa (98.55\%), and XLM-RoBERTa (98.76\%) are combined into a preliminary ensemble strategy as previously described. XLM (14.81\%) and XLNet (35.68\%) are omitted due to their low classification abilities. As noted, it was observed previously that the best score by a single model was RoBERTa which scored 98.96\% classification accuracy, and thus the main goal of the statistical ensemble classifier is to learn patterns that could possibly account for making up some of the 1.04\% of errors and correct for them. Initially, Table \ref{tab:ensemble-infogain} shows the information gain rankings of each predictor by 10 fold cross validation on the training set alone, interestingly BERT is ranked the highest with an information gain of 2.717 ($\pm$ 0.002). Following this, the results in Table \ref{tab:ensemble-results} show the results for multiple statistical methods of ensembling the predictions of the five Transformer models; all of the models with the exception of Gaussian Na\"{i}ve Bayes could outperform the best single Transformer model by an accuracy increase of at least +0.42 points. The two best models which achieved the same score were Logistic Regression and Random Forests, which when ensembling the predictions of the five transformers, could increase the accuracy by +0.63 points over RoBERTa and achieve an accuracy of 99.59\%. 

\begin{figure}
    \centering
    \includegraphics[scale=0.9]{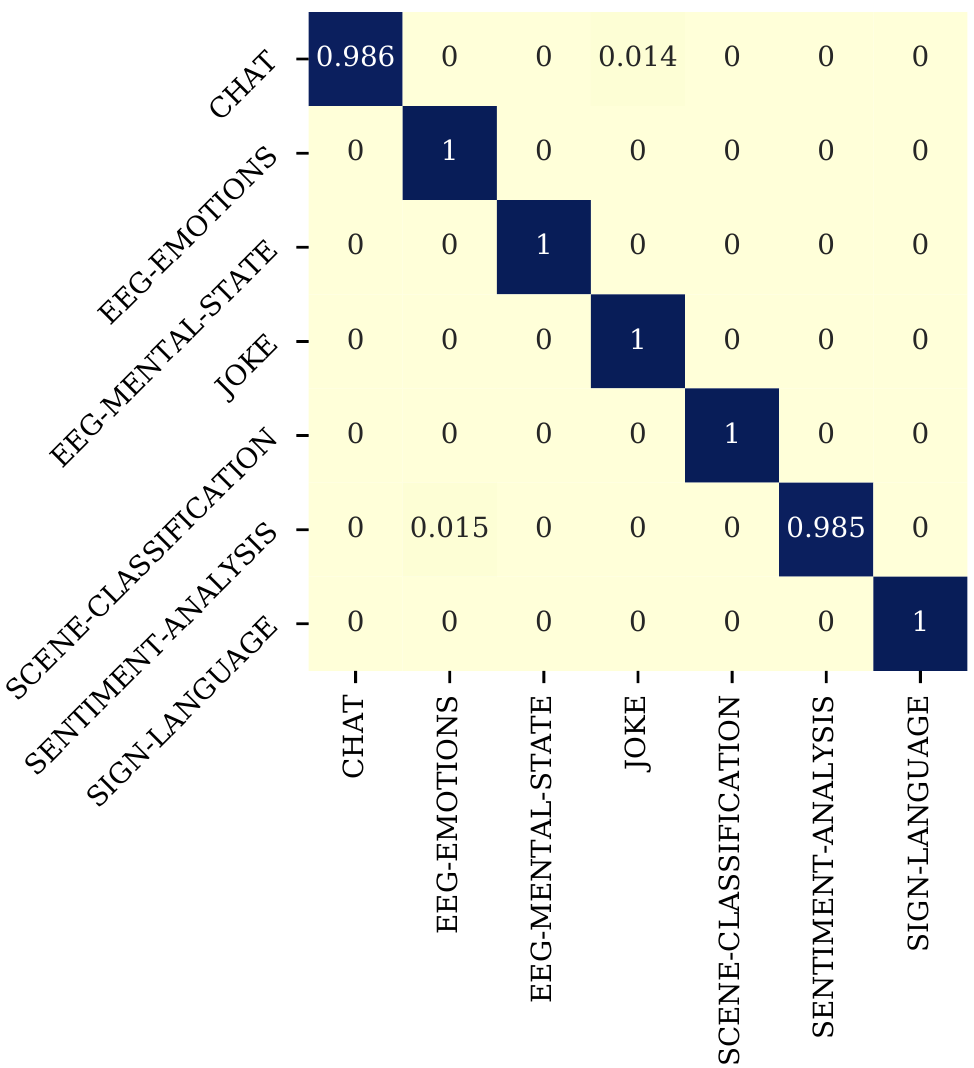}
    \caption{Normalised confusion matrix for the best ensemble methods of Logistic Regression and Random Forest (errors made by the two were identical).}
    \label{fig:ensemble-confusion}
\end{figure}

Finally, figure \ref{fig:ensemble-confusion} shows the confusion matrix for both the Logistic Regression and Random Forest methods of ensembling Transformer predictions since the errors made by both models were identical. Many of the errors have been mitigated through ensembling the transformer models, with minor confusion occuring between the 'CHAT' and 'JOKE' classes and the 'SENTIMENT ANALYSIS' and 'EEG-EMOTIONS' classes.

\section{Conclusion and Future Work}
\label{sec:conclusionfuturework}
The studies performed in this work have shown primarily that data augmentation through transformer-based paraphrasing via the T5 model have positively useful effects on many state-of-the-art language transformer-based classification models. BERT and DistilBERT, RoBERTa and DisilRoBERTa, XLM, XLM-RoBERTa, and XLNet all showed increases in learning performance when learning with augmented data from the training set when compared to learning only on the original data pre-augmentation. The best single model found was RoBERTa, which could classify human commands to an artificially intelligent system at a rate of 98.96\% accuracy, where errors were often due to ambiguity within human language. A statistical ensemble of the five best transformer models then led to an increase accuracy of 99.59\% when using either Logistic Regression or a Random Forest to process the output predictions of each transformer, utilising small differences between the models when trained on the dataset.

Although XLM did not perform well, the promising performance of XLM-RoBERTa showed that models trained on a task do not necessarily underperform on another different task given the general ability of lingual understanding. With this in mind, and given that the models are too complex to train simultaneously, it may be useful in future to consider the predictions of all trained models and form an ensemble through meta classifiers through statistical, deep learning, or further transformer approaches. A small vector input of predictions would allow for deeper decision making given the singular outputs of each transformer. Alternatively, a vector of inputs in addition to the original text may allow for deeper understanding behind why errors are made and allow for learned exceptions to overcome them. A preliminary ensemble of the five models that did not have weak scores showed that classification accuracy could be further increased by treating the outputs of each transformer model as attributes in themselves, for rules to be learnt from. The experiment was limited in that attribute selection was based solely on removing the two underperforming models; in future, exploration could be performed into attribute selection to fine-tune the number of models used as input. Additionally, only a predicted labels in the form of nominal attributes were used as input, whereas additional attributes such as probabilities of each output class could be utilised in order to provide more information for the statistical ensemble classifier.

\section{Ethics}
All users who answered the questionnaire agreed to the following statement:\\

The data collected from this form will remain completely anonymous and used for training a transformation-based chatbot. The more examples of a command or statement the bot can observe, the more accurate it will be at giving the correct response. The responses will be expanded by exploring paraphrases of answers and then further transformed by a model pre-trained on a large corpus of text and fine-tuned on the goal-based statements and requests given here.


\bibliographystyle{ieeetr}      
\bibliography{bibliography}   

\end{document}